\documentclass[letterpaper]{article}

\usepackage{algorithm}
\usepackage{algpseudocode}
\usepackage{glossaries}
\usepackage{csquotes}
\usepackage{todonotes}
\setuptodonotes{inline}
\usepackage{graphicx}
\usepackage{subcaption}
\usepackage{standalone}
\usepackage{dblfloatfix}
\usepackage{enumitem}
\setitemize{noitemsep,topsep=0pt,parsep=0pt,partopsep=0pt}
\setenumerate{noitemsep,topsep=0pt,parsep=0pt,partopsep=0pt}

\usepackage{natbib,alifeconf}  
\usepackage{url,hyperref}
\usepackage{cleveref}
\usepackage{booktabs}
\usepackage{lipsum}
\usepackage{siunitx}

\usepackage{tikz}
\usetikzlibrary{calc,positioning,arrows.meta,backgrounds}
\usepackage[rgb]{xcolor} 
\usepackage{amssymb}
\usepackage{amsmath}
\usepackage{wasysym}
\usepackage{graphicx}
\usepackage{calc}

\colorlet{c-prim}{blue!50!white}
\colorlet{c-prim-l}{c-prim!60}
\colorlet{c-sec}{red!50!white}
\colorlet{c-sec-l}{c-sec!60}
\tikzset{
    pics/text/.style args={#1,#2,#3}{
        code={
            \ifx#1w
                \node[anchor=east] (text) at (0,0) {#3};
                \draw[ultra thick, white,line cap=round] (0,0) -- ++ (0.2,0) -- (#2);
                \draw[thick,line cap=round] (0,0) -- ++ (0.2,0) -- (#2);
            \fi 
            \ifx#1e
                \node[anchor=west] (text) at (0,0) {#3};
                \draw[ultra thick, white,line cap=round] (0,0) -- ++ (-0.2,0) -- (#2);
                \draw[thick,line cap=round] (0,0) -- ++ (-0.2,0) -- (#2);
            \fi 
            \ifx#1n
                \node[anchor=south] (text) at (0,0) {#3};
                \draw[ultra thick, white,line cap=round] (0,0) -- ++ (0,-0.2) -- (#2);
                \draw[thick,line cap=round] (0,0) -- ++ (0,-0.2) -- (#2);
            \fi 
        }
    }
}

\newacronym[%
    longplural={cellular automata},
    shortplural={CA}
]{ca}{CA}{cellular automaton}

\newacronym{cnn}{CNN}{convolutional neural network}

\newacronym[%
    longplural={neural cellular automata},
    shortplural={NCA}
]{nca}{NCA}{neural cellular automaton}

\newacronym{uart}{UART}{Universal Asynchronous Receiver / Transmitter}
\newacronym{usb}{USB}{universal serial bus}
\newacronym{pio}{PIO}{Programmable Input Output}

\newacronym{led}{LED}{Light Emitting Diode}
\newacronym{pcb}{PCB}{Printed Circuit Board}
\newacronym{mse}{MSE}{mean squared error}

\title{A Rotation-Invariant Embedded Platform for (Neural) Cellular Automata}
\author{
    Dominik Woiwode\texorpdfstring{$^\dagger$}{}$^{1}$,
    Jakob Marten\texorpdfstring{$^\dagger$}{}$^{2}$, \and
    Bodo Rosenhahn$^{1}$ \\
    \texorpdfstring{$^\dagger$}{}~Contributed equally\\
    \mbox{}\\
    $^1$Institute of Information Processing, Leibniz University Hannover, Germany\\
    \{woiwode, rosenhahn\}@tnt.uni-hannover.de\\[3pt]
    $^2$Institute of Microelectronic Systems, Leibniz University Hannover, Germany \\
    marten@ims.uni-hannover.de
} 

\makeatletter
\newcommand\newtag[2]{\phantomsection #1\def\@currentlabel{#1}\label{#2}}  
\newcommand\myhiddenlabel[2]{\phantomsection \def\@currentlabel{#1}\label{#2}}  
\makeatother

\begin{document}

\maketitle

\begin{abstract}
    This paper presents a rotation-invariant embedded platform for simulating (neural) cellular automata (NCA) in modular robotic systems.
    Inspired by previous work on physical NCA, we introduce key innovations that overcome limitations in prior hardware designs.
    Our platform features a symmetric, modular structure, enabling seamless connections between cells regardless of orientation.
    Additionally, each cell is battery-powered, allowing it to operate independently and retain its state even when disconnected from the collective.
    To demonstrate the platform's applicability, we present a novel rotation-invariant NCA model for isotropic shape classification.
    The proposed system provides a robust foundation for exploring the physical realization of NCA, with potential applications in distributed robotic systems and self-organizing structures.
    Our implementation, including hardware, software code, a simulator, and a video, is openly shared at:
    \url{https://github.com/dwoiwode/embedded_nca}
\end{abstract}

\section{Introduction}



In nature, various phenomena, such as crystal formations and zebra stripes, can be described by \glspl{ca} \citep{shah_coupling_2022, gravan_evolving_2004}.
\Glspl{ca} are used for modeling discrete and dynamic systems.
They consist of simple cells that change their state simultaneously, following a uniform update rule applied to all cells.
This rule is constrained in a way that allows only local interactions with neighboring cells.
Despite this, life-like structures can emerge on 2D grids with relatively simple rules as in Conway's \enquote{Game of Life} \citep{gardner_mathematical_1970}.
Identifying update rules that produce emergent behavior is a difficult process, often carried out by hand in the past.
\Glspl{nca} seek to address this by learning the update rule through a neural network \citep{mordvintsev_growing_2020}.
They keep the local property of \glspl{ca} (e.g.\ a $3\times 3$ neighborhood on a 2D grid) and use stochastic gradient descent to optimize the update rule regarding a given target pattern.

Realizing software simulations in the physical domain often exposes gaps between modeled behavior and real-world dynamics.
In modular robotic systems, this presents a unique challenge, as the absence of global communication restricts information exchange to strictly local interactions.
Our work is closely related to the inspiring idea by \cite{walker_physical_2022} to bring \glspl{nca} into the physical world.
However, several assumptions in their paper lead to limitations that are addressed by our proposed hardware.
While prior work employs male and female header pins to ensure stable connections, this approach enforces a fixed orientation, which our design aims to overcome.
The proposed hardware is symmetric, allowing any side to connect seamlessly with any side of another cell.
The previous paper supplies power via cable, which introduces two key limitations: all active cells draw current through a single connection, leading to increased load issues, and modules cannot be rearranged while powered. 
In contrast, the proposed implementation utilizes battery-powered cells, enabling them to retain their state even when disconnected from the collective.
Since each unit operates independently, they can be assembled into swarms of virtually any size.

\begin{figure}[!t]
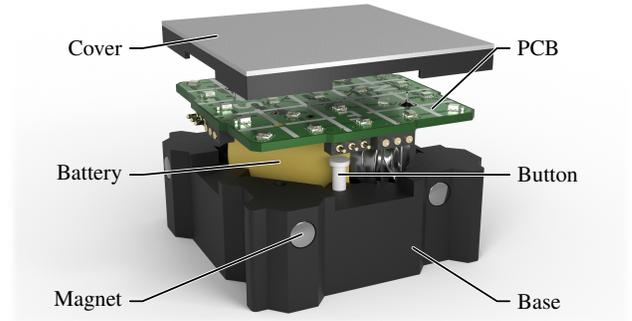

    \centering
    \includestandalone[width=\linewidth]{fig/hardware_design/stack}
    \caption{Structural breakdown of a single cell’s hardware components.}
    \label{fig:hardware_design:stack}
\end{figure}

To summarize, the \textbf{core contributions} of our work include:
\begin{itemize}
    \item We developed a portable, square-shaped robotic module that operates independently of its horizontal orientation and communicates with its four adjacent neighbors (see \nameref{sec:hardware}).
    \item We present a novel rotation-invariant \gls{nca} capable of classifying shapes regardless of their orientation (see \nameref{sec:software:nca}).    
    \item We analyze the effect of our rotation-invariant training methods and show their performance enhancements (see \nameref{sec:experiments}).
\end{itemize}

\section{Related Work}
A \gls{ca} is a computational model with a set of \enquote{cells} on a $d$-dimensional grid that update their state over time based on their local neighbors using the same update rule \citep{neumann_self-reproducing_1966}.
A cell state $s$ can range from a single boolean variable to a high-dimensional vector.
A classic example of a \gls{ca} is Conway's \enquote{Game of Life} \cite{gardner_mathematical_1970}, illustrating the emergence of complexity from simple local rules on a two-dimensional grid.
Throughout the remainder of this work, \gls{ca} refers to a two-dimensional cellular automaton with a cell state $s\in\mathbb{R}^c$.

A \acrfull{nca} is a special type of \gls{ca} which was introduced by \cite{mordvintsev_growing_2020}.
It is characterized by the update rule which is represented by a neural network.
Each \textbf{cell state} of a \gls{nca} has $c$ different channels in which cell specific information can be encoded. 
The \gls{nca} can use certain channels arbitrarily as hidden channels, whereas others are reserved for interpretable purposes.
Common examples include a RGB color representation in the first 3 channels \citep{mordvintsev_growing_2020, niklasson_self-organising_2021, otte_generative_2021}, a classification output \citep{randazzo_self-classifying_2020}, segmentation masks \citep{sandler_image_2020} or even gene encodings \citep{stovold_neural_2023}.

The cellular update rule of a \gls{nca} is split in two parts.
First, a \textbf{cell perception} is computed by applying various convolutional kernels.
The kernels operate channel-wise, and their outputs are concatenated to form the perception vector $P$.
While \cref{fig:perception:kernels} shows the kernels applied in our setup, a typical \gls{nca} allows for arbitrary local kernels.
Once the cell perception is computed, the \textbf{state update} $\Delta s$ is determined solely based on the perceptual information.
This process involves a relatively small neural network, typically containing fewer than 10,000 parameters, which makes them a perfect fit for small embedded hardware.
More details on \gls{nca}, based on our architectural design, are provided in section \nameref{sec:software:nca}.

Since the influential paper by \citeauthor{mordvintsev_growing_2020} in \citeyear{mordvintsev_growing_2020}, this subject has been the focus of many papers.
In recent years, more than 50 publications have been produced, which are aggregated and continuously updated in a repository by \cite{Woiwode_Awesome_Neural_Cellular_2025}.
Next, we present works which are relevant for this paper.
\cite{oliveira_approach_2008} published an approach to use \gls{ca} for 2D pattern recognition.
\cite{randazzo_self-classifying_2020} build on this idea and proposed an \gls{nca} which is capable of recognizing handwritten digits \citep{lecun_gradient-based_1998}.
The task of each cell in the \gls{nca} is to locally predict the global digit encoded by the structure.
Through repeated local exchanges of state, cells are able to indirectly communicate with distant parts and agree on a unified decision.
\citeauthor{walker_physical_2022} applied this concept in \citeyear{walker_physical_2022} in a real-world setting by constructing modular robotic devices that emulate \glspl{nca} to classify their global shape.
Hardware limitations required replacing the commonly used $3\times 3$ Moore neighborhood with the more constrained Von Neumann neighborhood, reducing each cell's accessible information~\citep{neumann_self-reproducing_1966}.
Similar to other \glspl{nca} approaches, their system depends on robots being aligned in a specific direction.

Later studies introduce a concept where each cell is given an individual orientation \citep{mordvintsev_growing_2022, randazzo_growing_2023}.
This concept is also used in our hardware implementation and further details will be presented in section \enquote{\nameref{subsec:software:nca:special}}.



\section{Hardware Design}
\label{sec:hardware}

The Kilobots, presented by \citet{rubenstein_kilobot_2012}, set a new standard for developing scalable and affordable robot swarms.
In this section, we introduce the proposed hardware architecture, which draws inspiration from \cite{walker_physical_2022} but further enables distinctive novel features such as persistent state, diverse rotations, and enhanced output capabilities.
Therefore, the hardware must meet three primary requirements:
\begin{enumerate}
    \item Each cell must be capable of communicating its current state to its four neighboring cells in any of the four rotational positions.
    \item Each cell should be able to display output via an array of colored \glspl{led}.
    \item Each cell must be powered by a battery to enable seamless disconnection and reconnection with other cells without losing its state.
\end{enumerate}
The first requirement has two major consequences for the design.
Firstly, a cell needs to be fitted with a genderless connector.
\citeauthor{walker_physical_2022} use pin headers for north/east and sockets for south/west placed in the center of each edge.
This prevents the rotation of the cell.
Our design uses two connectors per edge -- one male and one female -- which results in an interface independent of the cell's rotation.
To further simplify the handling of the cells, spring actuated pins and corresponding counter parts were chosen.
In total, each edge connects 6 pins, of which two pins each are used for power and ground.
The remaining pins are used to communicate between the two cells using the \gls{uart} protocol.

Secondly this directly requires that the microcontroller is able to communicate over four \gls{uart} channels at the same time.
Many microcontroller in the low-budget range only support one or two interfaces, forcing to use a software driven implementation for the remaining channels.
While technically possible, it reduced the amount of computation time available for the main calculations and additionally complicates the firmware design.
In our case we selected the RP2350 microcontroller by Raspberry Pi~Ltd~\citep{raspberrypipico} as it is both low-budget and can support more than four hardware-supported \gls{uart} channels.

\begin{figure}
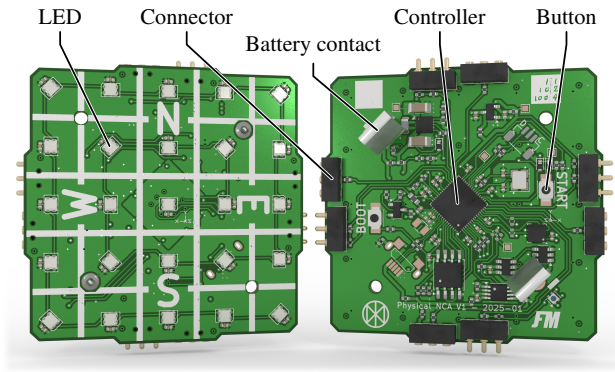

    \centering
    \includestandalone[width=\linewidth]{fig/hardware_design/pcb}
    \caption{The top (left) and bottom side (right) of the custom PCB.}
    \label{fig:hardware_design:pcb}
\end{figure}

To satisfy our second requirement, we chose to implement an array of 25 \glspl{led} in a grid of $5\times5$ which is shown on the left side of \cref{fig:hardware_design:pcb}.
This enables many different output patterns and a straightforward output of the current cell state.
As \gls{led} component a WS2812B-2020 was selected since it has RGB color output, a small spatial footprint and allows controlling a whole chain by only one data pin.

To be able to move a cell freely without any outgoing cables, an additional battery circuitry was added, which sits diagonally at the center of the \gls{pcb} (see right side of \cref{fig:hardware_design:pcb}).
It stabilizes the battery voltage to usable supply voltages of \num{5} and \qty{3.3}{V} for all components and enables its charging via \gls{usb} or the edge connector.
To prevent any damage to the battery, a battery protection circuit is included.
To balance physical size, ease of maintenance and capacity a rechargeable battery in the CR123A form factor was selected ($\qty{34}{mm} \times \qty{16}{mm}\ \diameter, \qty{700}{mAh}$). 

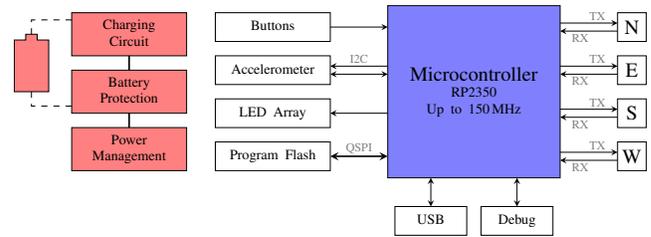
\begin{figure}
    \centering
        \begin{tikzpicture}[
        every node/.style={
            align=center,
            inner sep=0,
            outer sep=0,
        },
        battery/.pic={
            \draw[fill=c-sec] (0,0) -| (0.6,0.9) -- (0.4,0.9) |- (0.2,1) |- (0,0.9) --cycle; 
            \coordinate (-north) at (0.3,1);
            \coordinate (-south) at (0.3,0);
        }
    ]
        \node[fill=c-prim,draw, text width=3cm, minimum height=3cm] (rp2350) at (7,2.5) {Microcontroller\\\scriptsize RP2350\\[-0.5em]Up to 150\,MHz};

        \node[draw, text width=2cm, minimum height=0.5cm] (buttons) at (3.5,3.625) {\scriptsize\strut{}Buttons};
        \node[draw, text width=2cm, minimum height=0.5cm] (accel) at (3.5,2.875) {\scriptsize\strut{}Accelerometer};
        \node[draw, text width=2cm, minimum height=0.5cm] (led) at (3.5,2.125) {\scriptsize\strut{}LED Array};
        \node[draw, text width=2cm, minimum height=0.5cm] (flash) at (3.5,1.375) {\scriptsize\strut{}Program Flash};
        
        \node[draw, text width=0.5cm, minimum height=0.5cm] (conn0) at (9.75,1.375) {W};
        \node[draw, text width=0.5cm, minimum height=0.5cm] (conn1) at (9.75,2.125) {S};
        \node[draw, text width=0.5cm, minimum height=0.5cm] (conn2) at (9.75,2.875) {E};
        \node[draw, text width=0.5cm, minimum height=0.5cm] (conn3) at (9.75,3.625) {N};

        \node[draw, text width=1.25cm, minimum height=0.5cm] (usb) at (6.25,0.25) {\scriptsize\strut{}USB};
        \node[draw, text width=1.25cm, minimum height=0.5cm] (debug) at (7.75,0.25) {\scriptsize\strut{}Debug};

        \foreach \n in {conn0,conn1,conn2,conn3}{
            \draw[stealth-,transform canvas={yshift=0.07cm}] (\n) -- (\n -| rp2350.east) node[gray, above=0.05cm,pos=0.35] {\tiny TX};
            \draw[-stealth,transform canvas={yshift=-0.07cm}] (\n) -- (\n -| rp2350.east) node[gray, below=0.05cm,pos=0.65] {\tiny RX};
        }

        \draw[stealth-,transform canvas={yshift=0.07cm}] (accel) -- (accel -| rp2350.west) node[gray, above=0.05cm,pos=0.5] {\tiny I2C};
        \draw[stealth-stealth,transform canvas={yshift=-0.07cm}] (accel) -- (accel -| rp2350.west);
        \draw[stealth-] (led) -- (led -| rp2350.west);
        \draw[-stealth] (buttons) -- (buttons -| rp2350.west);
        
        \draw[stealth-stealth] (usb) -- (usb |- rp2350.south);
        \draw[stealth-stealth] (debug) -- (debug |- rp2350.south);
        
        \draw[stealth-stealth,thick] (flash) -- (flash -| rp2350.west) node[gray, above=0.05cm,pos=0.5] {\tiny QSPI};
 
        \node[fill=c-sec,draw, text width=2cm, minimum height=0.75cm,font=\scriptsize] (charger) at (1,3.5) {\strut{}Charging\\\strut{}Circuit};
        \node[fill=c-sec,draw, text width=2cm, minimum height=0.75cm,font=\scriptsize] (protection) at (1,2.5) {\strut{}Battery\\\strut{}Protection};
        \node[fill=c-sec,draw, text width=2cm, minimum height=0.75cm,font=\scriptsize] (power) at (1,1.5) {\strut{}Power\\\strut{}Management};

        \draw (-1,2.5) pic (battery) {battery};
        
        \draw[dashed] (battery-north) -- ++ (0,0.25) -- ($(charger.west)+(0,0.25)$);
        \draw[dashed] (battery-south) -- ++ (0,-0.25) -- ($(protection.west)+(0,-0.25)$);
        \draw[thick] (charger) -- (protection);
        \draw[thick] (protection) -- (power);
    \end{tikzpicture}
    \caption{Different components used in the hardware design.}
    \label{fig:hardware_design:block}
\end{figure}

The final design of the custom 4-layer \gls{pcb} as presented in \cref{fig:hardware_design:pcb} measures \qty{49}{mm}$\times$\qty{49}{mm}. 
The top side (left) only contains the \gls{led} array; the bottom side houses any other components, i.e. the connectors, the microcontroller, buttons and an 3-axis accelerometer for user interaction, an \gls{usb} and debug connector, the battery contacts, and the power and battery circuitry.
\Cref{fig:hardware_design:block} presents the interaction of the different components.

Beyond the \gls{pcb}, the hardware of a cell consists of four 3D-printed parts as well as eight magnets allowing a quick and stable connection of multiple cells.
The whole stack is displayed in \cref{fig:hardware_design:stack}.
A 3D-printed base houses the battery and magnets.
The \gls{pcb} is placed on the top.
To reach the buttons, two extension pins are inserted through the base.
A translucent cover protects the electronics and improves the perceptibility by scattering of the emitted light.

\subsection{Firmware}
\label{subsec:firmware}

The main purpose of the implemented microcontroller firmware lies in three tasks:
The communication with neighboring cells and therefore receiving and sending state, the calculation of the next state based on the current and received state and the display of the output.
The firmware is implemented in C using the \enquote{PICO C SDK}~\citep{raspberrypipico-sdk}.

For the first task, the four bidirectional ports are used.
Each port is connected to one edge connector and uses the \gls{uart} protocol.
As the RP2350 only has two hardware \gls{uart} interfaces, two \gls{pio} units of the microcontroller where used; one for transmitting data and one for receiving the incoming data.
The \gls{pio} units provide four in- or outputs each and can be used to emulate different transfer protocols. 
This flexibility also give the possibility to adapt existing protocols beyond their typical configuration options.
In our case, we chose to implement a \gls{uart} protocol with a word with of \qty{32}{bit}, simplifying the transmission of single precision floating point numbers used for state representation.
A baud rate of \qty{115200}{kBd} was selected.
The transfer is initiated by moving the data to be sent to the corresponding \gls{pio} data queues.
On the receiving site, the data is deserialized and placed into the \gls{pio}'s receiver queues.
An interrupt is issued to allow further processing by the main core.

\begin{figure}
    \centering
        \colorlet{colorheader}{c-sec}
    \colorlet{colorrtensor}{c-prim}
    \colorlet{colorwtensor}{c-prim-l}
    \colorlet{colorop}{c-sec-l}
    \begin{tikzpicture}[
        mem/.style = {
            inner sep = 1,
            outer sep = 0,
            minimum height = 0.5cm,
            text width = 3.5cm,
            align = left,
            draw = black,
            anchor = west,
            font = \ttfamily
        },
        tensor/.style = {
            text width = 3cm
        }
    ]
        \node at (1.75,0.5) {\bfseries Program};
    
        \node[mem, fill=colorheader] at (0,0) {Header};
        
        \node[mem, tensor, fill=colorrtensor] (r1) at (0.5,-0.5) {R1: 0x0000};
        \node[mem, tensor, align=center, fill=colorrtensor] at (0.5,-1) {\dots};
        \node[mem, tensor, fill=colorrtensor] (r5) at (0.5,-1.5) {R5: 0x0204};
        \node[rotate=90] at (0.25,-1) {\footnotesize Read only};
        
        \node[mem, tensor, fill=colorwtensor] (w6) at (0.5,-2) {W6: 0x0000};
        \node[mem, tensor, align=center, fill=colorwtensor] at (0.5,-2.5) {\dots};
        \node[mem, tensor, fill=colorwtensor] (w9) at (0.5,-3) {W9: 0x0102};
        \node[rotate=90] at (0.25,-2.5) {\footnotesize Writable};

        \node[mem, tensor, fill=colorop] at (0.5,-3.5) {OP1: 1 + 2 $\rightarrow$ 6};

        \node[mem, tensor, fill=colorrtensor, align=center] (r1mem) at (0.5,-4) {\dots}; 
        \node[mem, tensor, fill=colorrtensor, align=center] at (0.5,-4.5) {\dots}; 
        \node[mem, tensor, fill=colorrtensor, align=center] (r5mem) at (0.5,-5) {\dots};

        \draw[thick, -stealth, dashed] (r1) -- ++ (1.75,0) |- (r1mem);
        \draw[thick, -stealth, dashed] (r5) -- ++ (2,0) |- (r5mem);

        \node at (6,0.5) {\bfseries Engine};
        
        \node at (6, 0) {Current State};
        \node[mem, tensor, fill=colorwtensor] (in) at (4.5,-0.5) {W0: 0x0000};

        \foreach \y/\s in {0/Own,1/North,2/East,3/South,4/West}{
            \draw(8.5,0-\y*0.25+0.125) rectangle ++ (1.25,-0.25);
            \node[anchor=west] at (8.5+0.05,0-\y*0.25) {\scriptsize \s};
        }
        
        \draw[dashed, gray] (in.north east) -- (8.5,-0+0.125);
        \draw[dashed, gray] (in.south east) -- (8.5,-0+0.125-5*0.25);
        
        \node at (6, -1.5) {Buffer};
        \node[mem, tensor, fill=colorwtensor, align=center] (w6mem) at (4.5,-2) {\dots}; 
        \node[mem, tensor, fill=colorwtensor, align=center] at (4.5,-2.5) {\dots}; 
        \node[mem, tensor, fill=colorwtensor, align=center] (w9mem) at (4.5,-3) {\dots};
        
        \draw[thick, -stealth, dotted] (w6) -- (w6mem);
        \draw[thick, -stealth, dotted] (w9) -- (w9mem);
        
        \node at (6, -4) {Output};
        \node[mem, tensor, fill=colorwtensor] (out) at (4.5,-4.5) {W255: 0x0000};

        \foreach \x in {0,...,4}{
            \foreach \y in {0,...,4}{
                \draw (8.5+\x*0.25,-4-\y*0.25+0.125) rectangle ++ (0.25,-0.25);
            }
        }
        \draw[fill=red] (8.5+1*0.25,-4-1*0.25+0.125) rectangle ++ (0.25,-0.25);
        \draw[fill=green] (8.5+3*0.25,-4-1*0.25+0.125) rectangle ++ (0.25,-0.25);
        \draw[fill=blue] (8.5+1*0.25,-4-3*0.25+0.125) rectangle ++ (0.25,-0.25);

        \draw[dashed, gray] (out.north east) -- (8.5,-4+0.125);
        \draw[dashed, gray] (out.south east) -- (8.5,-4+0.125-5*0.25);
        
    \end{tikzpicture}
    \caption{Program and data structure and its interaction with the engine.}
    \label{fig:hardware_design:program}
\end{figure}
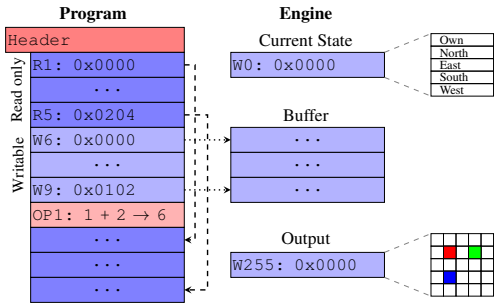

Within the second step, the calculation based on the current state of the cell and its neighbors is performed.
Therefore, all states are combined to a single tensor and fed to a configurable calculation engine.
Inspired by similar frameworks as TensorFlow Lite, a program is represented by a header, a set of tensors and an operations list.
Its structure is presented in \cref{fig:hardware_design:program}.

The header contains mandatory information as the version of the model, the size of the cell state $c$, the number of tensors and operations and delays used to control the timing behavior.
The set of tensors is represented by a list of pointer and can be split into two distinct types:
On the on hand a tensor can be immutable (read only=R).
Those tensors are used to provided values e.g. weights needed by the performed operations.
Their content is stored directly in the program and cannot be changed. 
On the other hand there are modifiable tensors used to store intermediate results (writable=W).
Their contents are stored within a global buffer area and referenced by the pointer value.
Each tensor has a \qty{8}{bit}, where 0 is reserved for the current state and 255 for the output tensor.
The output tensor has the dimension $25\times3$ where each \gls{led} has three channels used to represent red, green, and blue.
The operation list contains operation descriptors, which are processed linearly from start to end in each update cycle. 
Each descriptor starts with a operation identifier which is used to select the correct operation.
Currently, the engine supports 11 different operations listed in \cref{tab:hardware_design:operations}.
It is also possible to add further operations by extending the engine to support other calculation steps.
It is followed by a operation specific amount of source and destination tensor identifiers or constant values.
For example the operation \textit{ADD} takes two source and one destination tensor identifier as well as the length of the tensors.
During program generation is must be assured that all tensors are at least as long as specified in the operations as those constraints are not checked during calculation.   

In the last step, the written output is display via the \gls{led} array on the top.
The third available \gls{pio} is used to generate the control signals for the \textit{WS2812B} \glspl{led}.
This way the values can directly be shifted in an output queue and are automatically transferred to the \glspl{led}.

\begin{table}
    \centering
    \small
    \begin{tabular}{l|l|l}
        Code & Name & Description \\\hline
        \texttt{0x00} & NOP & No operation \\
        \texttt{0x01} & ADD & Add two tensors \\
        \texttt{0x02} & MAT\_MUL & Matrix multiplication on two tensors \\
        \texttt{0x03} & RELU & Apply relu on tensor \\
        \texttt{0x04} & FILL & Fill tensor with value \\
        \texttt{0x05} & MAX & Calculate maximum over tensor \\
        \texttt{0x06} & SOFTMAX & Calculate softmax over tensor \\
        \texttt{0x07} & STEP & Apply step function to tensor \\
        \texttt{0x08} & MUL & Multiplication on two tensors \\
        \texttt{0x0B} & FILL\_RAND & Fill tensor with random value \\
        \texttt{0x0C} & ARG\_MAX & Calculate argmax over tensor  
    \end{tabular}
    \caption{List of implemented operations.}
    \label{tab:hardware_design:operations}
\end{table}

In addition to the three main tasks, the firmware handles input data of the connected accelerometer.
Currently, the values are only used to detect if the cell is flipped upside-down to initiate a power off.
In future work, it is planned to enable access to those value from the calculation engine to include them in a program.
Also, a debug interface is implemented to transfer the current cell state and performance parameters via the \gls{usb} interface.

The \gls{pcb} layout, microcontroller firmware, simulator and 3D model files are open source and available in our code.

\subsection{Simulator}
\label{subsec:simulator}
To perform experiments quickly without the need to compile the firmware and apply it to all hardware cells, a simulator running in the browser was developed.
It can be used to simulate and debug different programs.
To ensure simulation accuracy, the firmware calculation engine is embedded as WebAssembly into the application.

\begin{figure}
    \centering
    \begin{subfigure}{0.48\linewidth}
        \centering
        \includegraphics[width=\linewidth]{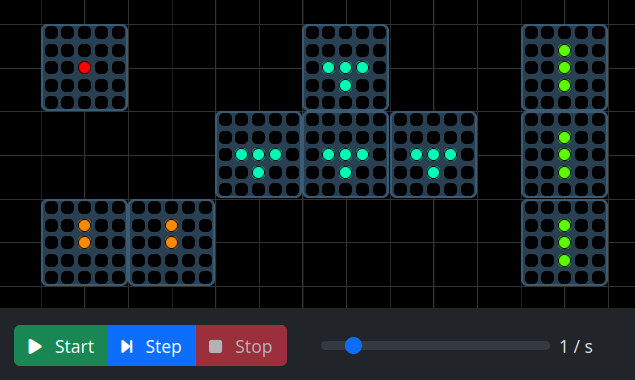}
        \caption{2D world with cells}
        \label{fig:hardware_design:simulator:a}
    \end{subfigure}
    \begin{subfigure}{0.48\linewidth}
        \centering
        \includegraphics[width=\linewidth]{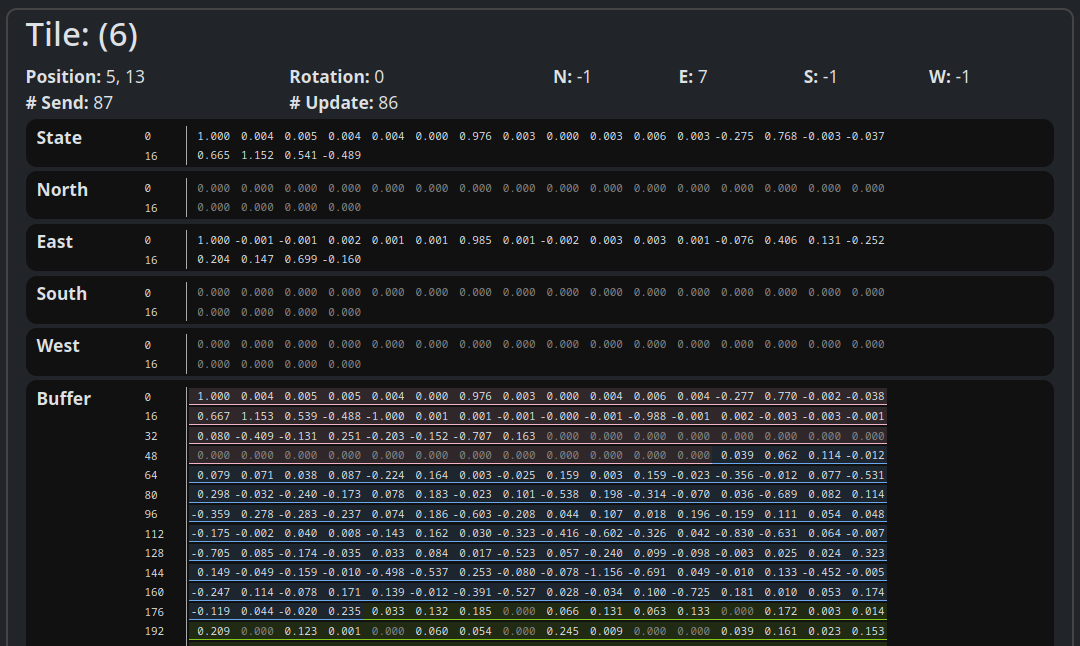}
        \caption{Selected cell information}
        \label{fig:hardware_design:simulator:b}
    \end{subfigure}
    \begin{subfigure}{0.48\linewidth}
        \centering
        \includegraphics[width=\linewidth]{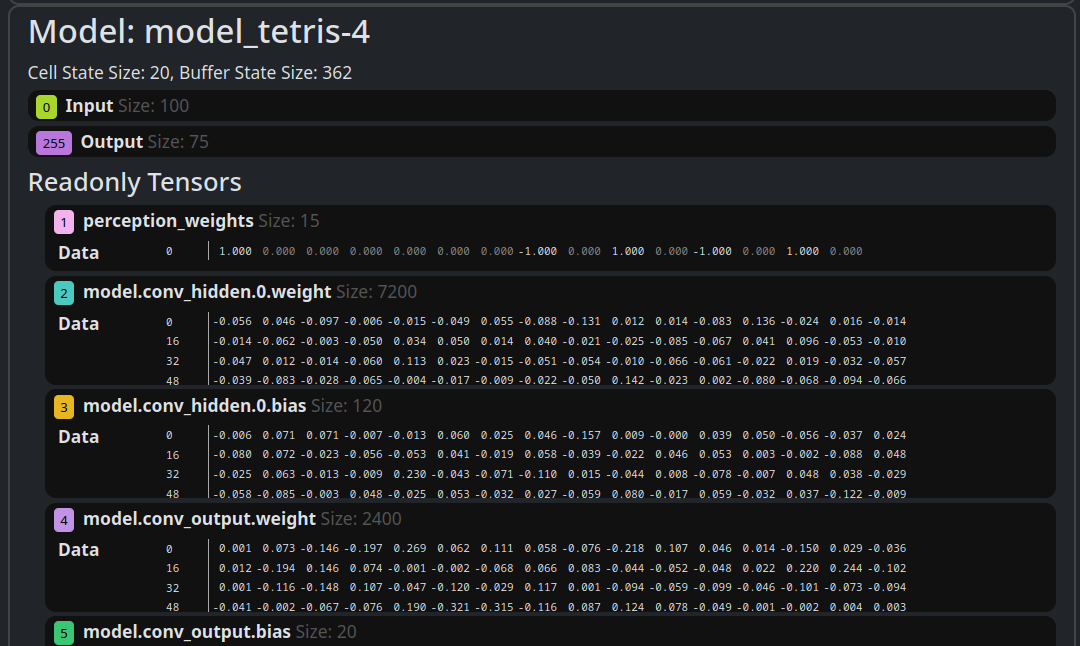}
        \caption{Model weights}
        \label{fig:hardware_design:simulator:c}
    \end{subfigure}
    \begin{subfigure}{0.48\linewidth}
        \centering
        \includegraphics[width=\linewidth]{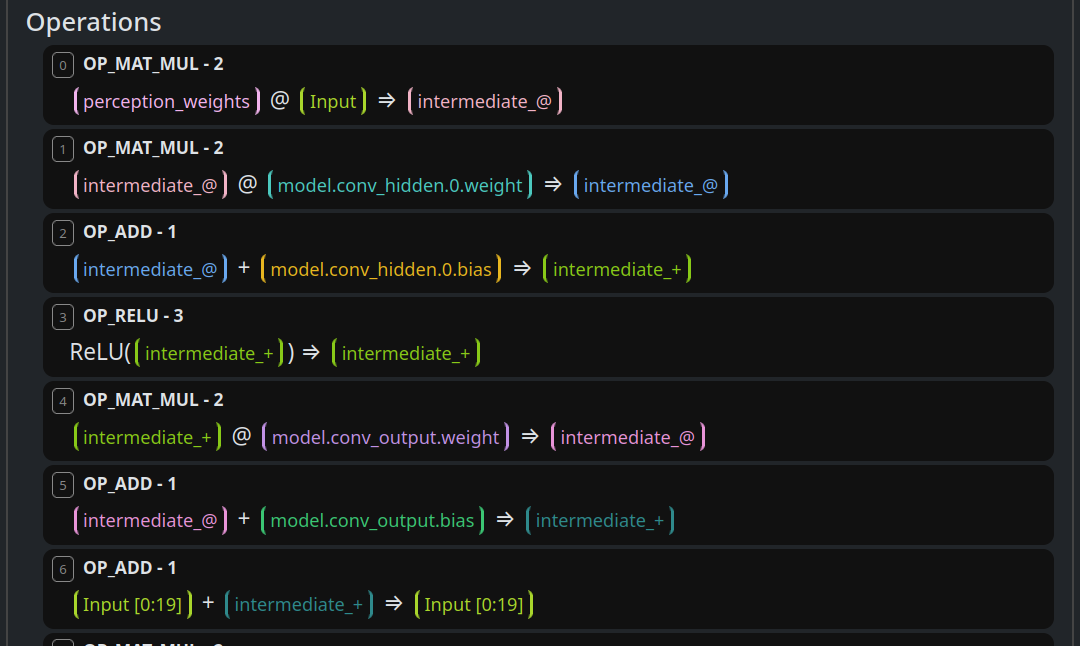}
        \caption{Operations list (model)}
        \label{fig:hardware_design:simulator:d}
    \end{subfigure}
    \caption{Different core parts of the simulation environment.}
    \label{fig:hardware_design:simulator}
\end{figure}

Examples images of the simulator are displayed in \cref{fig:hardware_design:simulator}.
The main part of the simulator are cells, displaying their current output, that can be moved and rotated (\ref{fig:hardware_design:simulator:a}).
A simulation can be controlled by different buttons as \textit{start}, \textit{step} and \textit{stop}.
Furthermore, it is possible to select other (precompiled) models, upload own models and chose different simulation techniques.
The state of a selected cell can be inspected within an information overview (\ref{fig:hardware_design:simulator:b}).
These include the current state, the received neighbor states and the buffer containing the content of intermediate tensors as well as the output.
The simulator also shows data on the used model e.g. the data of the immutable included tensors (\ref{fig:hardware_design:simulator:c}).
Additionally a list of operation is displayed (\ref{fig:hardware_design:simulator:d}).

\section{Training a physical NCA}
\label{sec:software:nca}
To minimize the required capabilities of the embedded hardware, we separated the training loop from the firmware of the robots.
This section will focus on the training procedure, which is implemented using a modular PyTorch framework.
We present specific adaptations necessary for achieving rotation-invariant training.
The framework and the training code is made open source and can be found in our code repository.

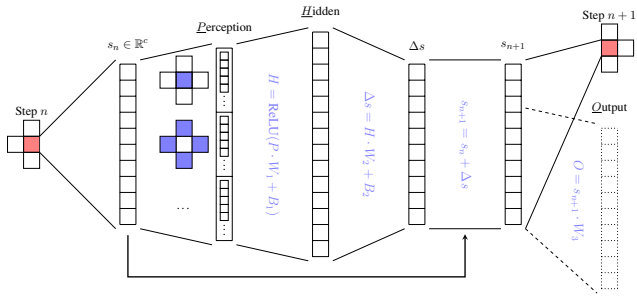
\begin{figure}[!htp]
    \centering
        \begin{tikzpicture}[
        font=\large,
        cellsleft/.pic = {
            \node[inner sep=0,text width=5mm, minimum height=5mm,draw,fill=c-sec] (-center) at (0,0) {};
            \node[inner sep=0,text width=5mm, minimum height=5mm,draw] at (-0.5,0) {};
            \node[inner sep=0,text width=5mm, minimum height=5mm,draw] at (0,+0.5) {};
            \node[inner sep=0,text width=5mm, minimum height=5mm,draw] at (0,-0.5) {};
        },
        cellsright/.pic = {
            \node[inner sep=0,text width=5mm, minimum height=5mm,draw,fill=c-sec] (-center) at (0,0) {};
            \node[inner sep=0,text width=5mm, minimum height=5mm,draw] at (+0.5,0) {};
            \node[inner sep=0,text width=5mm, minimum height=5mm,draw] at (0,+0.5) {};
            \node[inner sep=0,text width=5mm, minimum height=5mm,draw] at (0,-0.5) {};
        },
        neighbours1/.pic = {
            \node[inner sep=0,text width=5mm, minimum height=5mm,draw,fill=c-prim] (-center) at (0,0) {};
            \node[inner sep=0,text width=5mm, minimum height=5mm,draw] at (+0.5,0) {};
            \node[inner sep=0,text width=5mm, minimum height=5mm,draw] at (-0.5,0) {};
            \node[inner sep=0,text width=5mm, minimum height=5mm,draw] at (0,+0.5) {};
            \node[inner sep=0,text width=5mm, minimum height=5mm,draw] at (0,-0.5) {};
        },
        neighbours2/.pic = {
            \node[inner sep=0,text width=5mm, minimum height=5mm,draw] (-center) at (0,0) {};
            \node[inner sep=0,text width=5mm, minimum height=5mm,draw,fill=c-prim] at (+0.5,0) {};
            \node[inner sep=0,text width=5mm, minimum height=5mm,draw,fill=c-prim] at (-0.5,0) {};
            \node[inner sep=0,text width=5mm, minimum height=5mm,draw,fill=c-prim] at (0,+0.5) {};
            \node[inner sep=0,text width=5mm, minimum height=5mm,draw,fill=c-prim] at (0,-0.5) {};
        }
    ]
        \node at (-3,1) {Step $n$};
        \draw (-3,0) pic (startcell) {cellsleft};
    
        \node (state) at (0,0) {\tikz{\draw (0,0) grid[step=0.5] (0.5,5)}}; 
        \node[above=2mm of state] {$s_n \in \mathbb{R}^c$};
        \draw (1.75,2) pic {neighbours1};
        \draw (1.75,0) pic {neighbours2};
        \node at (1.75,-2) {\dots};

        \node (perception) at (3,0) {\tikz{
            \draw (0,0) grid[step=0.5,ystep=2](0.5,6);
            \foreach \p in {0,2,4}{
                \foreach \i in {2,...,6}{
                    \draw (0.125,0.125+\i*0.25+\p) rectangle ++ (0.25,0.25);
                }
                \node at (0.25, \p+0.45) {\vdots};
            }
        }};
        \node[above=2mm of perception] {\underline{$P$}erception};

        \node[c-prim,rotate=-90] at (4.5,0) {\Large{}$H=\text{ReLU}(P \cdot W_1 + B_1) $};
        
        \node (h) at (6,0) {\tikz{\draw (0,0) grid[step=0.5] (0.5,7)}}; 
        \node[above=2mm of h] {\underline{$H$}idden};

        \node[c-prim,rotate=-90] at (7.5,0) {\Large{}$\Delta s = H \cdot W_2 + B_2$};

        \node (deltastate) at (9,0) {\tikz{\draw (0,0) grid[step=0.5] (0.5,5)}}; 
        \node[above=2mm of deltastate] {$\Delta s$};

        \node (nextstate) at (12,0) {\tikz{\draw (0,0) grid[step=0.5] (0.5,5)}}; 
        \node[above=2mm of nextstate] {$s_{n+1}$};

        \node (classification) at (15,-2) {\tikz{\draw[dotted, thick] (0,0) grid[step=0.5] (0.5,5)}}; 
        \node[above=2mm of classification] {\underline{$O$}utput};

        \node at (15,4) {Step $n+1$};
        \draw (15,3) pic (nextstep) {cellsright};

        \node[c-prim,rotate=-90] at (10.5,0) {\Large{}$s_{n+1}=s_n + \Delta s $};
        
        \node[c-prim,rotate=-90] at (14,-1.75) {\Large{}$O=s_{n+1} \cdot W_3 $};
        
        \fill[white,opacity=0.7] (startcell-center.north east) -- (state.north west) -- (state.south west) -- (startcell-center.south east) -- cycle;
        \draw (startcell-center.north east) -- (state.north west);
        \draw (startcell-center.south east) -- (state.south west);

        \draw (state.north east) -- (perception.north west);
        \draw (state.south east) -- (perception.south west);
        
        \draw (perception.north east) -- (h.north west);
        \draw (perception.south east) -- (h.south west);
        
        \draw (h.north east) -- (deltastate.north west);
        \draw (h.south east) -- (deltastate.south west);
        
        \draw (deltastate.north east) -- (nextstate.north west);
        \draw (deltastate.south east) -- (nextstate.south west);

        \draw[dashed] ($(nextstate.north east)+(0,-1.5)$) -- (classification.north west);
        \draw[dashed] (nextstate.south east) -- (classification.south west);

        \draw (nextstate.north east) -- (nextstep-center.north west);
        \draw (nextstate.south east) -- (nextstep-center.south west);
        
        \draw[-stealth,very thick] ($(state.south)+(0,-0.2)$) |- ($(state.south)+(0,-1.5)$) -| (10.5, -2.8);
    \end{tikzpicture}
    \caption{Overview of our \gls{nca} model. The use of a task head to generate task-specific outputs is optional and not implemented in all models.}
    \label{fig:software:nca_overview}
\end{figure}

An overview of our \gls{nca} model can be seen in \cref{fig:software:nca_overview}.
Our \gls{ca} is based on a 2D grid of cells, where each cell can see the state of their directly adjacent cells (\enquote{\textit{4-neighborhood}}) \citep{neumann_self-reproducing_1966}).

\begin{figure}
    \centering
        \begin{tikzpicture}[
        pics/colornode/.style args={#1,#2}{
            code = {
                \ifnum#1=1
                    \node[fill=c-prim] {#1};
                \else
                    \ifnum#1=-1
                        \node[fill=c-sec] {#1};
                    \else
                        \ifnum#2=1
                            \node {#1};
                        \else
                            \node[text=gray] {#1};
                        \fi
                    \fi
                \fi
            }
        },
        pics/kernel/.style args={#1,#2,#3,#4,#5,#6,#7,#8,#9}{
            code = {
                \matrix [nodes={draw,fill=white,minimum size=7mm},ampersand replacement=\&]
                {
                    \draw pic {colornode={#1,0}}; \& \draw pic {colornode={#2,1}}; \& \draw pic {colornode={#3,0}}; \\
                    \draw pic {colornode={#4,1}}; \& \draw pic {colornode={#5,1}}; \& \draw pic {colornode={#6,1}}; \\
                    \draw pic {colornode={#7,0}}; \& \draw pic {colornode={#8,1}}; \& \draw pic {colornode={#9,0}}; \\
                };
            }   
        }
    ]
        \draw (0,0) pic (identity) {kernel={0,0,0,0,1,0,0,0,0}};
        \node at (0,-1.5) {\strut{}Identity};

        \draw (2.5,0) pic (gradx) {kernel={0,0,0,-1,0,1,0,0,0}};
        \node at (2.5,-1.5) {\strut{}Gradient X};

        \draw (5,0) pic (grady) {kernel={0,-1,0,0,0,0,0,1,0}};
        \node at (5,-1.5) {\strut{}Gradient Y};

        \draw (7.5,0) pic (neumann) {kernel={0,1,0,1,0,1,0,1,0}};
        \node at (7.5,-1.5) {\strut{}Von Neumann};
    \end{tikzpicture}
    \caption{The perception kernels used for our \gls{nca} model.}
    \label{fig:perception:kernels}
\end{figure}
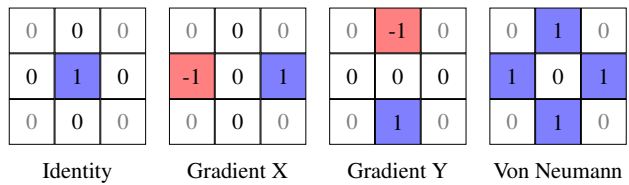
The perception is built upon a combination of different kernels $K$ that are adapted to our 4-neighborhood, as shown in \cref{fig:perception:kernels}.
\enquote{Gradient X} and \enquote{Gradient Y} are simple gradient filters similar to Sobel filter \citep{sobel_33_1973}.
The \enquote{Von Neumann} is a isotropic kernel that sums all neighboring values.
The \enquote{Identity} is simply the own state of the cell.
These kernels are applied independently per channel and the results are subsequently concatenated to form a perception vector $P\in \mathbb{R}^{c\cdot n_k}$, where $n_k$ is the number of kernels applied.
\begin{align}
    P = \text{concat} \left( K_1 \overset{\text{cw}}{\otimes} s,\; K_2 \overset{\text{cw}}{\otimes} s,\; \dots,\; K_{n_k} \overset{\text{cw}}{\otimes} s \right),
    \label{eq:perception}
\end{align}
where $\overset{\text{cw}}{\otimes}$ denotes a channel-wise convolution.
In this context, $s$ refers to the state configuration of the entire \gls{ca}, where the kernels inherently account for the local 4-neighborhood interactions.
Constant zero-padding is used at the boundary regions, as this accurately represents inactive cells.

We calculate the next state for a cell following a simple update rule
\begin{align}
    H &= \text{ReLU}(P \cdot W_1 + B_1),\\
    \Delta s &= H \cdot W_2 + B_2,\\
    s_{n+1} &= s_{n} + \Delta s,
    \label{eq:next_step}
\end{align}
where $W_i$ and $B_i$ are the learnable weights and bias of the $i$-th layer of the neural network respectively and ReLU is the Rectified Linear Unit activation function \citep{nair_rectified_2010}.

A training iteration starts with a seed state $s_0$ and applies the \gls{nca} update rule $T$ times, where $T$ is randomly sampled between $10$ and $40$ for our experiments.
Instead of the commonly used cross-entropy loss for classification, a \gls{mse} against a one-hot encoded class label $l$ is used, as it yielded more stable training results \cite{walker_physical_2022,randazzo_self-classifying_2020}.

In contrast to a classical \gls{nca}, we added an optional task specific head layer.
Consequently, the number of channels needed is less constrained by the task.
For example, classifying 29 categories typically requires at least 29 channels, yet one-hot encoding causes most channels to remain near zero, conveying minimal information.
With a classification layer, the \gls{nca} is able to learn a latent space representation of the classification.
To keep the spirit of the original \gls{nca}, only a part of the final state is forwarded to this layer.
This preserves the benefit of hidden channels, which can serve purposes beyond the primary task, such as encoding spatial relationships with neighboring cells.
The task head layer only has to be used when the \gls{nca} is evaluated and has no direct impact on the further state.
We apply the loss defined in \cref{eq:loss} and train the model using gradient descent and the Adam optimizer \citep{kingma_adam_2014}.

\subsection{Hardware-aware training strategies}
\label{subsec:software:nca:special}
    While the preceding section outlines the (mostly) standard training procedure for \gls{nca}, the characteristics and constraints of our robot hardware impose specific considerations that must be integrated into the training process.

    Setting the first channel of each cell always to $1$ ensures that the cells are able to sense each other. 
    Otherwise, lacking neighbors can be indistinguishable from having a neighboring cell in a \enquote{dead} state, defined as a zero vector.
    We use this channel to apply a (modified\footnote{The original Alive-Masking also takes neighboring cells into account. We limit this to the own cell state.}) \enquote{\textbf{Alive-Masking}} to the cells \citep{mordvintsev_growing_2020,randazzo_self-classifying_2020}.
    \begin{align}
        \Delta s' = \begin{cases}
            \Delta s & \text{ if cell is active}\\
            0 & \text{ otherwise}
        \end{cases}
    \end{align}

    As our robots are designed to be \textbf{rotation invariant}, we also incorporate this aspect in our training.
    With our hardware, the cells can be physically rotated to achieve a different orientation.
    During training we add an extra channel to our \gls{nca} that represents the orientation angle $\theta$ of this cell.
    This angle cannot be directly perceived by any cell, including itself, and it is inherently accommodated within the hardware design due to the physical capability to rotate the cells, rather than explicitly encoded.
    We then apply a 2D rotation matrix to rotate our gradient perceptions $P_x$ and $P_y$ at $\theta$ to get $P'_x$ and $P'_y$ using following formula
    \begin{align}
        \begin{bmatrix}
            P'_x \\ P'_y 
        \end{bmatrix}
            = \begin{bmatrix}
                \cos\theta & - \sin\theta\\
                \sin\theta & \cos\theta\\
            \end{bmatrix}
            \begin{bmatrix}
                P_x \\ P_y
            \end{bmatrix}.
    \end{align}

\begin{figure*}[!b]
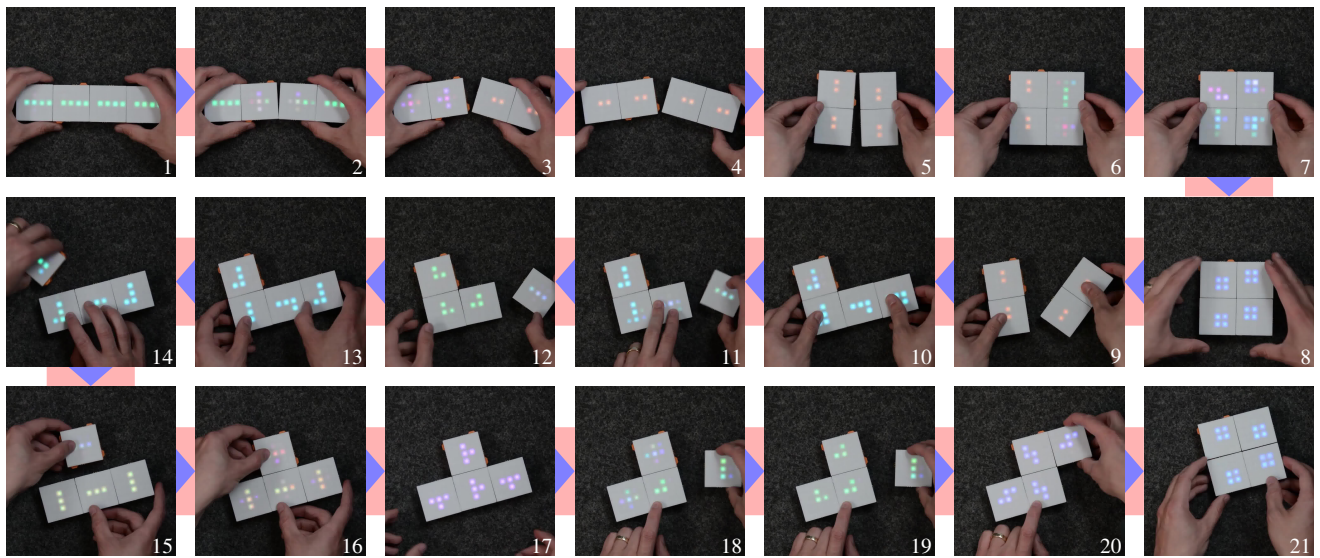

  \centering
  \includestandalone[width=\linewidth]{fig/experiments/video-sequence}
  \caption{Example sequence of images from a video demonstrating the shape recognition. Some frames (e.g. 6 and 7) show intermediate states before classifying the correct shape again (frame 8). Frames 10--13 demonstrate the ability to rotate a tile. The full video can be seen on our project page.}
  \label{fig:video:images}
\end{figure*}

    The same concept already has been used by \cite{mordvintsev_growing_2020} on a global scale and by \cite{randazzo_growing_2023} for each cell.
    Their approaches are modified by discretizing $\theta$ into four predefined rotations: $0^{\circ}, 90^{\circ}, 180^{\circ}, 270^{\circ}$.
    $\theta$ is fixed during one training iteration and cannot be changed by the \gls{nca} itself.
    
    While other implementations focus on getting a correct result at a specific simulation step \citep{walker_physical_2022}, we want to achieve a stable configuration over a long time.
    We therefore make use of a \textbf{training pool} \citep{mordvintsev_growing_2020}.
    A pool of samples is initialized with corresponding seed and target configurations.
    At the start of one training iteration, a batch of samples is drawn from this pool.
    Then, this batch is propagated through the \gls{nca} and evaluated as described above.
    At the end of each training iteration, the selected batch is added back to the sample pool, enabling longer episode durations over time.
    In this process, $k_\text{new}$ random samples are substituted with new seeds and targets, preventing the unlearning of previous iterations.
    Furthermore, $k_\text{replaced}$ random samples are altered to new targets, where only the channel indicating the living cells and $\theta$ are modified, but the overall cell state keeps intact.
    This modification was essential for enabling the models to learn to recognize new shapes.
    Otherwise, they remained stuck in a fixed state, despite changes in configuration.
    
    We simulate an \textbf{asynchronicity} of cells by adding a random dropout mask to $\Delta s$.
    It has been shown that this makes the model more robust to pertubations in the state of the \gls{ca} \citep{niklasson_asynchronicity_2021}.
    We also add a small normal distributed noise ($\sigma = 2\cdot 10^{-2}$) to each update that is not affected by the dropout, which further helps the model to recover from small pertubations \citep{randazzo_self-classifying_2020}.
    
    After training a model, the corresponding network structure and model weights have to be converted into a format that is recognizable by the firmware.
    During this step most modifications explained in this section are stripped, as they are only relevant during training.
    Because each cell determines its next state independently, the overall neural network is minimal and can be summarized by equations (\ref{eq:perception})--(\ref{eq:next_step}).

\section{Experiments}
\label{sec:experiments}

We show the capabilities of our robotic models in two different experiments.
In the first experiment, we evaluate our hardware using a hand-crafted model.
For this we use a task called \enquote{firefly-synchronization}, where different cells have to agree on a common clock frequency.

In the second experiment, we reproduce the classic shape classification for simple digits using a \gls{nca}. 
We also extend this experiment using rotation invariant polyominoes up to a degree of 5.
Our results show that cells can be added, removed or shifted in the configuration and also keep their state while being alone.

\subsection{Firefly Synchronization}
To demonstrate the capabilities of our hardware and firmware, we choose the \enquote{firefly-synchronization} task.
The synchronization of pulse-coupled oscillating cells is a widely researched topic \citep{mirollo_synchronization_1990,brandner_firefly_2016,macau_modeling_2019}.
Each cell has an internal clock which can be adjusted by interacting with their local neighbors to achieve synchrony.
When the internal clock reaches its period time of $1$ it resets and the cell emits a pulse which can be perceived by other cells.
Their goal is to reach a global synchronization with decentralized actions and is inspired by the natural behavior of fireflies flashing in unison.
We adapt and implement this algorithm for our cell-like robots using the \enquote{Von Neumann} kernel shown in \cref{fig:perception:kernels}.
\Cref{alg:firefly} shows a pseudocode of our implementation.
\begin{algorithm}
\caption{Pseudocode of a single cell update step.}
\begin{algorithmic}
\Procedure{CellUpdate}{~}
    \State \texttt{state} $\gets$ \texttt{state} + \texttt{k}
    \If{\texttt{neighbor\_flash\_detected()}}
        \State \texttt{state} $\gets$ \texttt{state} + (\texttt{k} $\cdot$ \texttt{state}) + random\_noise
    \EndIf
    \If{\texttt{state} $\geq$ 1}
        \State \Call{flash}{~}
        \State \texttt{state} $\gets$ 0
    \EndIf
\EndProcedure
\end{algorithmic}
\label{alg:firefly}
\end{algorithm}

\begin{figure}
    \centering
    \includegraphics[width=\linewidth]{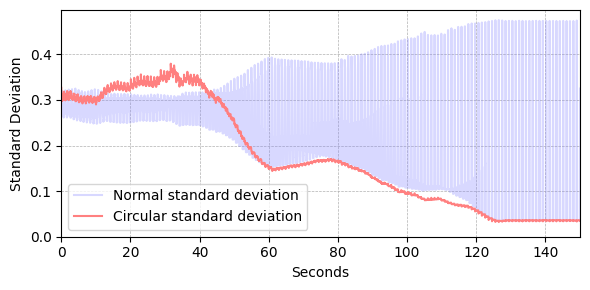}
    \caption{Standard deviation of the cell phases over time. Initially, the phases are randomly distributed. The simulation involves 29 cells arranged in a circle with a diameter of 5.}
    \label{fig:eval:firefly}
\end{figure}

\begin{figure}
    \centering
    \includegraphics[width=0.45\linewidth]{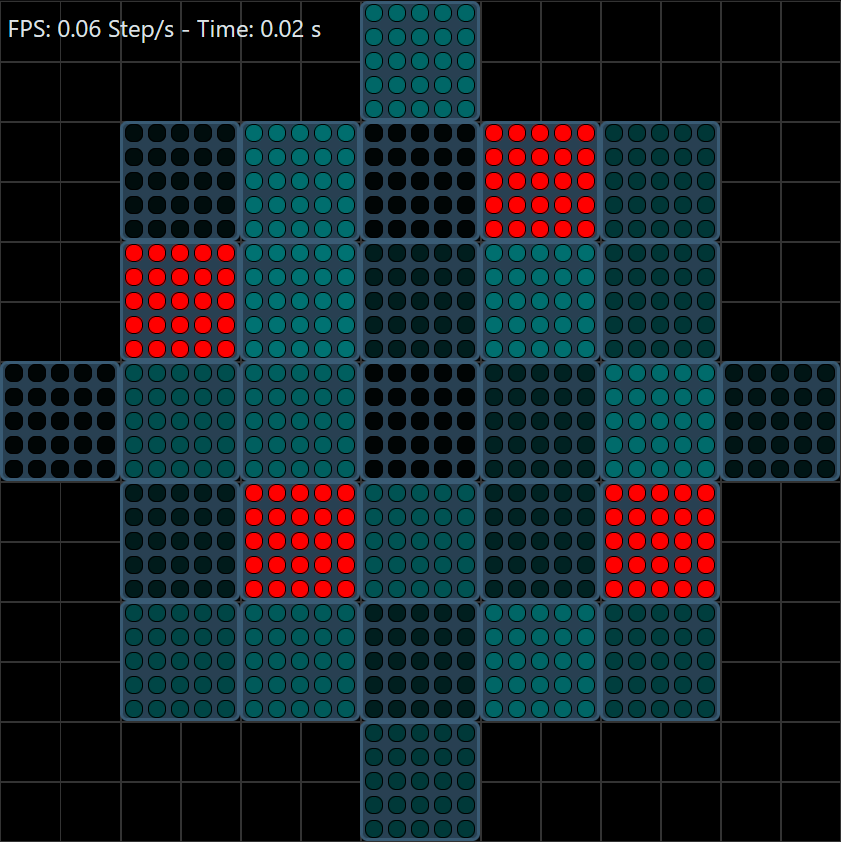}
    \quad
    \includegraphics[width=0.45\linewidth]{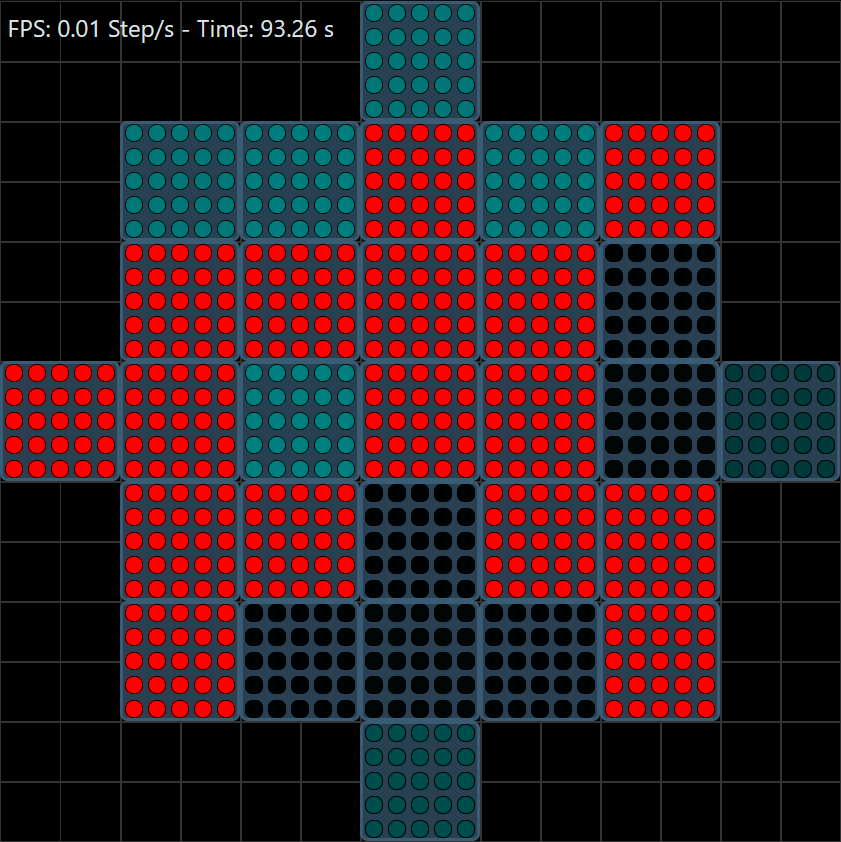}
    \caption{Simulator screenshots illustrating the firefly synchronization experiment setup. Cell brightness represents phase value; red cells indicate flashes. Left: initial random cell initialization. Right: setup after  \qty{90}{s}.}
    \label{fig:firefly:setup}
\end{figure}

We compute a circular standard deviation $\sigma_\circ$ from directional statistics for all $n$ cells with a phase angle $\varphi$:
\begin{align}
    R &= \frac{1}{n}\sqrt{\left(\sum_{i=1}^n \sin \varphi_i\right)^2 + \left(\sum_{i=1}^n \cos \varphi_i\right)^2}\\
    \sigma_\circ &= \frac{\sqrt{-2 \ln R}}{2 \pi}
\end{align}
In contrast to a normal standard deviation, a circular standard deviation takes into account that the $\varphi$ wraps around so that $0\equiv 1 \mod 1$. 
As seen in \cref{fig:eval:firefly} the cells find a common phase after roughly 2 minutes in our simulation.
This behavior can also be observed qualitatively as seen in \cref{fig:firefly:setup}.
This time could be adjusted by changing the phase shift parameters $k$.

\subsection{Shape Classification}
\begin{figure}
    \centering
    \begin{subfigure}{0.48\linewidth}
        \centering
            \definecolor{colorzero}{hsb}{0,1,1}
    \definecolor{colorone}{hsb}{0.09,1,1}
    \definecolor{colortwo}{hsb}{0.18,1,1}
    \definecolor{colorthree}{hsb}{0.27.09,1,1}
    \definecolor{colorfour}{hsb}{0.36,1,1}
    \definecolor{colorfive}{hsb}{0.45,1,1}
    \definecolor{colorsix}{hsb}{0.55,1,1}
    \definecolor{colorseven}{hsb}{0.64,1,1}
    \definecolor{coloreight}{hsb}{0.73,1,1}
    \definecolor{colornine}{hsb}{0.82,1,1}
    \begin{tikzpicture}[
        pics/block/.style args={#1}{
            code = {
                \draw[fill=#1,thick] (-0.25,-0.25) rectangle ++ (0.5,0.5); 
            }
        }
    ]
        \begin{scope}[shift={(0,0)}]
            \draw (0,0) pic {block=colorzero};
            \draw (0,-0.5) pic {block=colorzero};
            \draw (0,-1) pic {block=colorzero};
            \draw (0,-1.5) pic {block=colorzero};
            \draw (0,-2) pic {block=colorzero};
            \draw (-1,0) pic {block=colorzero};
            \draw (-1,-0.5) pic {block=colorzero};
            \draw (-1,-1) pic {block=colorzero};
            \draw (-1,-1.5) pic {block=colorzero};
            \draw (-1,-2) pic {block=colorzero};
            \draw (-0.5,0) pic {block=colorzero};
            \draw (-0.5,-2) pic {block=colorzero};
        \end{scope}
        \begin{scope}[shift={(2,0)}]
            \draw (0,0) pic {block=colorone};
            \draw (0,-0.5) pic {block=colorone};
            \draw (0,-1) pic {block=colorone};
            \draw (0,-1.5) pic {block=colorone};
            \draw (0,-2) pic {block=colorone};
        \end{scope}
        
        \begin{scope}[shift={(4,0)}]
            \draw (0,0) pic {block=colortwo};
            \draw (0,-0.5) pic {block=colortwo};
            \draw (0,-1) pic {block=colortwo};
            \draw (0,-2) pic {block=colortwo};
            \draw (-1,0) pic {block=colortwo};
            \draw (-1,-1) pic {block=colortwo};
            \draw (-1,-1.5) pic {block=colortwo};
            \draw (-1,-2) pic {block=colortwo};
            \draw (-0.5,0) pic {block=colortwo};
            \draw (-0.5,-1) pic {block=colortwo};
            \draw (-0.5,-2) pic {block=colortwo};
        \end{scope}
        
        \begin{scope}[shift={(6,0)}]
            \draw (0,0) pic {block=colorthree};
            \draw (0,-0.5) pic {block=colorthree};
            \draw (0,-1) pic {block=colorthree};
            \draw (0,-1.5) pic {block=colorthree};
            \draw (0,-2) pic {block=colorthree};
            \draw (-1,0) pic {block=colorthree};
            \draw (-1,-1) pic {block=colorthree};
            \draw (-1,-2) pic {block=colorthree};
            \draw (-0.5,0) pic {block=colorthree};
            \draw (-0.5,-1) pic {block=colorthree};
            \draw (-0.5,-2) pic {block=colorthree};
        \end{scope}
        
        \begin{scope}[shift={(8,0)}]
            \draw (0,0) pic {block=colorfour};
            \draw (0,-0.5) pic {block=colorfour};
            \draw (0,-1) pic {block=colorfour};
            \draw (0,-1.5) pic {block=colorfour};
            \draw (0,-2) pic {block=colorfour};
            \draw (-1,0) pic {block=colorfour};
            \draw (-1,-0.5) pic {block=colorfour};
            \draw (-1,-1) pic {block=colorfour};
            \draw (-0.5,-1) pic {block=colorfour};
        \end{scope}
        
        \begin{scope}[shift={(0,-3)}]
            \draw (0,0) pic {block=colorfive};
            \draw (0,-1) pic {block=colorfive};
            \draw (0,-1.5) pic {block=colorfive};
            \draw (0,-2) pic {block=colorfive};
            \draw (-1,0) pic {block=colorfive};
            \draw (-1,-0.5) pic {block=colorfive};
            \draw (-1,-1) pic {block=colorfive};
            \draw (-1,-2) pic {block=colorfive};
            \draw (-0.5,0) pic {block=colorfive};
            \draw (-0.5,-1) pic {block=colorfive};
            \draw (-0.5,-2) pic {block=colorfive};
        \end{scope}
        
        \begin{scope}[shift={(2,-3)}]
            \draw (0,0) pic {block=colorsix};
            \draw (0,-1) pic {block=colorsix};
            \draw (0,-1.5) pic {block=colorsix};
            \draw (0,-2) pic {block=colorsix};
            \draw (-1,0) pic {block=colorsix};
            \draw (-1,-0.5) pic {block=colorsix};
            \draw (-1,-1) pic {block=colorsix};
            \draw (-1,-1.5) pic {block=colorsix};
            \draw (-1,-2) pic {block=colorsix};
            \draw (-0.5,0) pic {block=colorsix};
            \draw (-0.5,-1) pic {block=colorsix};
            \draw (-0.5,-2) pic {block=colorsix};
        \end{scope}
        
        \begin{scope}[shift={(4,-3)}]
            \draw (0,0) pic {block=colorseven};
            \draw (0,-0.5) pic {block=colorseven};
            \draw (0,-1) pic {block=colorseven};
            \draw (0,-1.5) pic {block=colorseven};
            \draw (0,-2) pic {block=colorseven};
            \draw (-1,0) pic {block=colorseven};
            \draw (-0.5,0) pic {block=colorseven};
        \end{scope}
        
        \begin{scope}[shift={(6,-3)}]
            \draw (0,0) pic {block=coloreight};
            \draw (0,-0.5) pic {block=coloreight};
            \draw (0,-1) pic {block=coloreight};
            \draw (0,-1.5) pic {block=coloreight};
            \draw (0,-2) pic {block=coloreight};
            \draw (-1,0) pic {block=coloreight};
            \draw (-1,-0.5) pic {block=coloreight};
            \draw (-1,-1) pic {block=coloreight};
            \draw (-1,-1.5) pic {block=coloreight};
            \draw (-1,-2) pic {block=coloreight};
            \draw (-0.5,0) pic {block=coloreight};
            \draw (-0.5,-1) pic {block=coloreight};
            \draw (-0.5,-2) pic {block=coloreight};
        \end{scope}
        
        \begin{scope}[shift={(8,-3)}]
            \draw (0,0) pic {block=colornine};
            \draw (0,-0.5) pic {block=colornine};
            \draw (0,-1) pic {block=colornine};
            \draw (0,-1.5) pic {block=colornine};
            \draw (0,-2) pic {block=colornine};
            \draw (-1,0) pic {block=colornine};
            \draw (-1,-0.5) pic {block=colornine};
            \draw (-1,-1) pic {block=colornine};
            \draw (-1,-2) pic {block=colornine};
            \draw (-0.5,0) pic {block=colornine};
            \draw (-0.5,-1) pic {block=colornine};
            \draw (-0.5,-2) pic {block=colornine};
        \end{scope}

    \end{tikzpicture}
        \caption{\textit{Digits} dataset. 
        }
        \label{fig:dataset:numbers}
    \end{subfigure}
    \hfill
    \begin{subfigure}{0.48\linewidth}
        \centering
            \definecolor{colorone}{hsb}{0,1,1}
    \definecolor{colortwo}{hsb}{0.09,1,1}
    \definecolor{colorthree}{hsb}{0.18,1,1}
    \definecolor{colorfour}{hsb}{0.27.09,1,1}
    \definecolor{colorfive}{hsb}{0.36,1,1}
    \definecolor{colorsix}{hsb}{0.45,1,1}
    \definecolor{colorseven}{hsb}{0.55,1,1}
    \definecolor{coloreight}{hsb}{0.64,1,1}
    \definecolor{colornine}{hsb}{0.73,1,1}
    \definecolor{colorten}{hsb}{0.82,1,1}
    \definecolor{coloreleven}{hsb}{0.9,1,1}
    \begin{tikzpicture}[
        pics/block/.style args={#1}{
            code = {
                \draw[fill=#1,thick] (-0.25,-0.25) rectangle ++ (0.5,0.5); 
            }
        }
    ]   
        \begin{scope}[shift={(0,0)}]
            \draw (0,0) pic {block=colorone};
        \end{scope}

        \begin{scope}[shift={(2,0)}]
            \draw (0,0) pic {block=colortwo};
            \draw (0,-0.5) pic {block=colortwo};
        \end{scope}

        \begin{scope}[shift={(4,0)}]
            \draw (0,0) pic {block=colorthree};
            \draw (0,-0.5) pic {block=colorthree};
            \draw (0,-1) pic {block=colorthree};
        \end{scope}

        \begin{scope}[shift={(6,0)}]
            \draw (0,0) pic {block=colorfour};
            \draw (0,-0.5) pic {block=colorfour};
            \draw (0.5,-0.5) pic {block=colorfour};
        \end{scope}

        \begin{scope}[shift={(8,0)}]
            \draw (0,0) pic {block=colorfive};
            \draw (0,-0.5) pic {block=colorfive};
            \draw (0,-1) pic {block=colorfive};
            \draw (0,-1.5) pic {block=colorfive};
        \end{scope}
        
        \begin{scope}[shift={(10,0)}]
            \draw (0,0) pic {block=colorsix};
            \draw (-0.5,-1) pic {block=colorsix};
            \draw (0,-0.5) pic {block=colorsix};
            \draw (0,-1) pic {block=colorsix};
        \end{scope}

        \begin{scope}[shift={(1,-2.5)}]
            \draw (0,0) pic {block=colorseven};
            \draw (0,-0.5) pic {block=colorseven};
            \draw (0,-1) pic {block=colorseven};
            \draw (0.5,-1) pic {block=colorseven};
        \end{scope}
        
        \begin{scope}[shift={(3,-2.5)}]
            \draw (0,0) pic {block=coloreight};
            \draw (0,-0.5) pic {block=coloreight};
            \draw (0.5,-0.5) pic {block=coloreight};
            \draw (0.5,0) pic {block=coloreight};
        \end{scope}
        
        \begin{scope}[shift={(5,-2.5)}]
            \draw (0,0) pic {block=colornine};
            \draw (-0.5,-0.5) pic {block=colornine};
            \draw (0,-0.5) pic {block=colornine};
            \draw (0.5,0) pic {block=colornine};
        \end{scope}

        \begin{scope}[shift={(7,-2.5)}]
            \draw (0,0) pic {block=colorten};
            \draw (-0.5,0) pic {block=colorten};
            \draw (0,-0.5) pic {block=colorten};
            \draw (0.5,0) pic {block=colorten};
        \end{scope}

        \begin{scope}[shift={(9,-2.5)}]
            \draw (0,0) pic {block=coloreleven};
            \draw (-0.5,0) pic {block=coloreleven};
            \draw (0,-0.5) pic {block=coloreleven};
            \draw (0.5,-0.5) pic {block=coloreleven};
        \end{scope}

    \end{tikzpicture}
        \caption{\textit{Polyomino-4} dataset.}
        \label{fig:dataset:polyominos}
    \end{subfigure}
    \caption{Datasets used for shape classification.}
    \label{fig:datasets}
\end{figure}
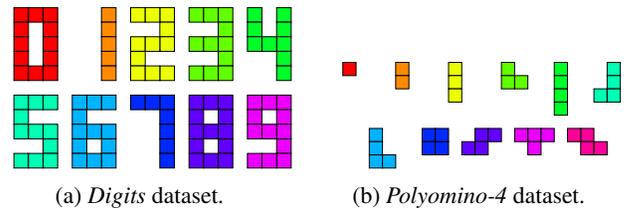

Inspired by \citet{randazzo_self-classifying_2020} and \citet{walker_physical_2022}, we test our robot model using a self-classifying shape detection task.
We create four different datasets for this task:
\begin{itemize}
    \item \textbf{digits}: Represents all digits from 0 to 9. This dataset is visualized in \cref{fig:dataset:numbers}.
    \item\textbf{digits-symmetric}: Represents only digits from 0 to 8, as \enquote{6} and \enquote{9} look the same when rotated by 180$^\circ$.
    \item\textbf{polyomino-4}: All different polyominoes up to a degree of 4 without accounting for the rotation of the shapes. This dataset is visualized in \cref{fig:dataset:polyominos}.
    \item\textbf{polyomino-5}: All different polyominoes up to a degree of 5 without accounting for the rotation of the shapes.
\end{itemize}


The architecture of the \gls{nca} is described in section \nameref{sec:software:nca}.
For the perception, we use the identity and two gradient kernels for X- and Y-direction which can be rotated during training as previously described.
One might be tempted to use an already isotropic kernel, but this approach is limited by the inability to distinguish certain shapes (e.g., both polyominoes of degree 3).
Therefore we use our rotation-invariant training described in section \nameref{subsec:software:nca:special}.

The final classification output $o(i)$ for a cell $i$ is either computed via the learned classification layer $W_C\in\mathbb{R}^{R\times C}$ applied to the state channels $s_n[:R]$, or directly taken from a subset of state channels.
\begin{align}
    o &= \begin{cases}
        s_n[:R] \cdot W_C & \text{ if classification layer is used}\\
        s_n[1:C+1] & \text{ otherwise}\\ 
    \end{cases}\\
    \mathcal{L} &= \frac{1}{N}\sum_{i=1}^{N}\left(o(i) - l\right)^2,
    \label{eq:loss}
\end{align}
where $C$ is the number of classes, $R$ is the number of channels used for the classification layer and $N$ is the number of active cells.

The classification output $o$ is used to compute a weighted sum of a predefined $5\times5$ visual representation of each class, which can be rendered on the \gls{led} screen. 
When the model is confident, $o$ approximates a one-hot vector, resulting in a clear class image; otherwise, a visual blend of multiple classes may appear (see e.g.\ frame 3 in \cref{fig:video:images}).

Per default, a batch size of 512, a sample pool size of 5120, 14 channels per cell state, 120 neurons for the hidden layer of the update function, and a classification layer which takes the first 10 channels as input is used.
A series of images from the resulting default model can be seen in \cref{fig:video:images}.

We evaluate the impact of the training methods by conducting several experiments.
Each experiment is repeated 5 times, reporting the mean and standard deviation.
Performance is measured using \textit{classification accuracy}, which verifies whether the argmax of the classification output matches the target label.

Firstly, the default training parameters are used to train a \gls{nca} for each dataset.
The classification remains stable 
As expected, the \texttt{polyomino-5} dataset is the hardest dataset, as it has the most different classes.
The results can be seen in \cref{tab:eval:datasets}.
\begin{figure}[!t]
    \centering
    \begin{subfigure}{\linewidth}
        \centering
        \includegraphics[width=\linewidth]{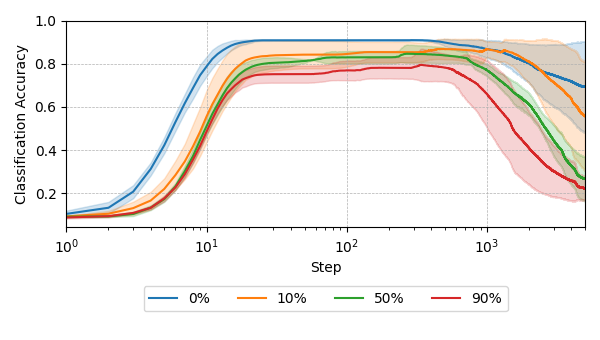}
        \caption{Classification performance during inference with unchanged targets. Note the logarithmic scale on the x-axis.}
        \label{subfig:tetris-4:target_replace_not_changing}
    \end{subfigure}
    
    \begin{subfigure}{\linewidth}
        \centering
        \includegraphics[width=\linewidth]{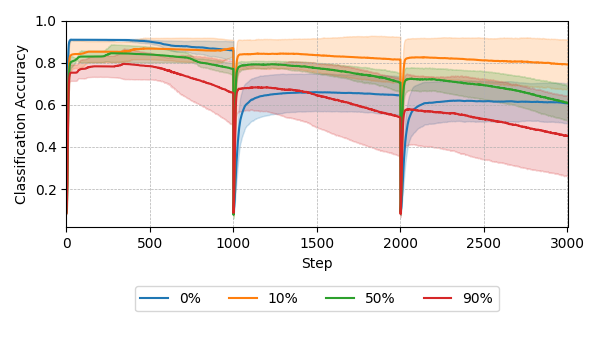}
        \caption{Classification performance during inference with target changes every 1000 steps.}
        \label{subfig:tetris-4:target_replace_changing}
    \end{subfigure}
    
    \caption{Long-term stability analysis on the \texttt{polyomino-4} dataset with different values for target replacement during training. Shaded areas indicate standard deviation.}
    \label{fig:tetris-4:target_replace}
\end{figure}

\begin{table}[!tp]
    \centering
    \begin{tabular}{l|ccc}
& \multicolumn{3}{c}{Inference Steps} \\Dataset & 50 & 100 & 150 \\
\hline
\texttt{digits (n)} & $0.90 \pm 0.01$ & $0.90 \pm 0.02$ & $0.90 \pm 0.02$ \\
\hline
\texttt{digits} & $0.81 \pm 0.09$ & $0.80 \pm 0.08$ & $0.80 \pm 0.08$ \\
\texttt{digits-sym} & $0.96 \pm 0.03$ & $0.95 \pm 0.03$ & $0.95 \pm 0.03$ \\
\texttt{polyomino-4} & $0.84 \pm 0.04$ & $0.83 \pm 0.04$ & $0.84 \pm 0.04$ \\
\texttt{polyomino-5} & $0.40 \pm 0.02$ & $0.40 \pm 0.02$ & $0.39 \pm 0.03$ \\
\end{tabular}
    \caption{Classification accuracy after several steps for different datasets.}
    \label{tab:eval:datasets}
\end{table}

We also investigate whether target mutation during training is essential.
\gls{nca} for \texttt{polyomino-4} are trained with target replacement rates of 0, 0.1, 0.5, and 0.9, and their long-term continuity is evaluated across 5000 iterations.
While \cref{subfig:tetris-4:target_replace_not_changing} indicates a negative impact of replacement, this is challenged by the next experiment with periodic target replacement every 1000 steps.
\Cref{subfig:tetris-4:target_replace_changing} suggests that changes in the cell configuration lead to even more stable classification results with a replacement rate of 0.1.




\section{Conclusion}
To summarize, we present an embedded hardware system that can emulate 2D \gls{ca}.
A central aspect of our design is its rotation-invariance, demonstrated through learning suitable \gls{nca} for shape classification.
Our work can serve as an effective educational tool to demonstrate the principles and dynamics of (neural) \gls{ca} to students.
To this end, our work could be extended by incorporating other \gls{nca} variants.
Additionally, our proposed output head could be leveraged to train more complex \gls{nca} models that generate structured outputs on the $5\times5$ \gls{led} matrix, enabling deeper insights into emergent behavior and learning dynamics.
This work is open to extension, and we encourage the community to contribute using our openly available hardware and software designs.


\section{Acknowledgements}
This work was supported by the Federal Ministry of Education and Research (BMBF), Germany, under the AI service center KISSKI (grant no. 01IS22093C), the MWK of Lower Saxony within Hybrint (VWZN4219) and the European Union  under grant agreement no. 101136006 – XTREME.

We used a large language model to refine and rephrase specific sentences.

\footnotesize
\bibliographystyle{apalike}
\bibliography{nca}
~

\end{document}